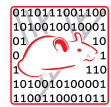

JOURNAL OF BIOMEDICAL SEMANTICS

RESEARCH  Open Access

# What's unusual in online disease outbreak news?

Nigel Collier[1,2]

Correspondence: collier@nii.ac.jp
[1]National Institute of Informatics, 2-1-2 Hitotsubashi, Chiyoda-ku, Tokyo 101-8430, Japan

## Abstract

**Background:** Accurate and timely detection of public health events of international concern is necessary to help support risk assessment and response and save lives. Novel event-based methods that use the World Wide Web as a signal source offer potential to extend health surveillance into areas where traditional indicator networks are lacking. In this paper we address the issue of systematically evaluating online health news to support automatic alerting using daily disease-country counts text mined from real world data using BioCaster. For 18 data sets produced by BioCaster, we compare 5 aberration detection algorithms (EARS C2, C3, W2, F-statistic and EWMA) for performance against expert moderated ProMED-mail postings.

**Results:** We report sensitivity, specificity, positive predictive value (PPV), negative predictive value (NPV), mean alerts/100 days and F1, at 95% confidence interval (CI) for 287 ProMED-mail postings on 18 outbreaks across 14 countries over a 366 day period. Results indicate that W2 had the best F1 with a slight benefit for day of week effect over C2. In drill down analysis we indicate issues arising from the granular choice of country-level modeling, sudden drops in reporting due to day of week effects and reporting bias. Automatic alerting has been implemented in BioCaster available from http://born.nii.ac.jp.

**Conclusions:** Online health news alerts have the potential to enhance manual analytical methods by increasing throughput, timeliness and detection rates. Systematic evaluation of health news aberrations is necessary to push forward our understanding of the complex relationship between news report volumes and case numbers and to select the best performing features and algorithms.

## Background

Recent public health emergencies of international concern (PHEICs), such as the ongoing A(H1N1) influenza pandemic, have highlighted the necessity for early warning systems that can detect events before they spread internationally. Strengthening health surveillance capacity is mandated by the revised International Health Regulations (IHR) [1] but efforts vary greatly from country to country according to resource availability and local priorities. For regions with weak indicator-based systems, event-based warning systems [2] using open media sources are increasingly seen as a viable and cost-effective alternative to strengthen global readiness. Where capacity already exists, indicator networks might also benefit from event-based data as a complementary source, e.g. in cutting down on manual reporting delays.

In the last decade a small but growing number of global health intelligence systems (GHISs) have exploited natural language technologies such as information retrieval, text classification and text mining, e.g. [3-8]. Whilst human expertise and judgement remain crucial, machine analysis is required to leverage the vast quantities of reports that require processing in real time each day. Upstream this can mean separating news into outbreak and non-outbreak classes. At the downstream end, it can involve making preliminary decisions on alerting in order to draw the expert's attention to reports that require detailed risk assessment. Free text analysis on hospital records such as chief complaint reports [9] has been previously used to support alerting, but to the best of our knowledge no detailed qualitative studies have been conducted to evaluate automatic health alerting with media sources. This study seeks to address this crucial gap [10] both from an end user perspective and to contribute toward the selection of outbreak detection algorithms. In such a complex task, we also look beneath aggregated accuracy statistics and examine in depth some of the challenge areas for text understanding.

Automatically detecting early signs of the unusual in news data requires the combination of sophisticated text mining to convert text to a structured event representation and aberration detection algorithms to discover whether the feature counts are exceptionally high. The minimum necessary requirement is to have the disease and country names properly identified. Operational systems such as BioCaster [4,11], which we use here as a test bed, also provide more sophisticated conceptual analysis. Examples include the spread of an outbreak over international borders, the infection of hospital workers or an accidental/deliberate release. Other similar freely available automatic systems include EpiSpider [3], HealthMap [5] and MedISys [7].

As with other aberration detection tasks, we need to be able to understand each event signal in a geotemporal context. Underlying seasonal factors, local and global trends as well as reporting bias are all factors in 'muddying the waters'. There is also a cultural and political dimension that cannot be ignored particularly when we consider distal indicators: the decision to close a school or a hospital ward for example may be standard procedure in one country but may signal infrastructure stress in another [12].

In this study we will focus only on the outbreak relationship automatically identified between disease and country. These by themselves are often surprisingly challenging to disambiguate. A single news report may contain country names for not only the location of the outbreak but also for the news organization's country, the responder's country, historical outbreak countries and those unrelated countries found in the headlines of linked reports. Disease names also provide their share of ambiguity with various forms that require normalization, for example: *swine flu*, *A(H1N1) influenza*, *H1N1 porcine influenza*, *Mexican influenza* and *pig flu* should all be normalized to the root term *A(H1N1) influenza*.

Despite a growing volume of literature, existing evaluations of GHISs (e.g. see [13] for a comparison) have up to now been fragmented and focused largely on component tasks such as text classification or on characteristics such as alerts per day, user accesses or system throughput. For example, Freifeld et al. [5] report automated topic classification accuracy of 84% overall on Google News and ProMED-mail alerts; Steinberger et al. [7] showed 88% PPV for alerting relevancy using a stratified approach to news volume variations. Mawudeku and Blench [14] report that between July 1998 and August 2001 56% of WHO verified outbreaks were initially found by the Global Public Health Intelligence Network (GPHIN) system; Doan et al. [15] report a study using a supervised machine learning for text classification with 93.5% accuracy; and Kawazoe et al. [16] highlight the ongoing challenge to classify ambiguous biomedical names by reporting F1 scores for term recognition. Whilst such evaluations provide valuable insights for system developers, they are not easy to compare, do not provide an obvious path for systematic improvement or fully characterize the nature of the early detection task. For these reasons we have chosen to focus here on alerting accuracy and an openly available 'gold standard'.

As an approximate gold standard benchmark we perform a side by side analysis against the curated postings from ProMED-mail [17], a human network of expert volunteers operating 24/7 as an official program of the International Society for Infectious Diseases. ProMED-mail volunteers monitor global media reports and other sources to provide Internet-based reports on biological and chemical hazards that affect humans, animals and plants. Each report undergoes a staged reviewing procedure by specialist moderators before it is either rejected or appears on the subscriber mailing list and Web site. Using ProMED-mail reports as a benchmark has several advantages: (a) they are openly and freely available and can serve as common evaluation criteria, (b) ProMED-mail has a high level of global coverage on most infectious diseases that public health agencies are interested in detecting, (c) the monitoring objectives of GHIS users seem to be broadly the same as those of ProMED-mail, indicated by the fact that the reports are widely used by the same community of end users and by health intelligence systems currently in use, and (d) the timeliness of reporting by Pro-MED-mail has been observed in many cases to be better than WHO outbreak verification reports [18].

Aberration detection algorithms are designed to look for dramatic upward changing patterns over time. From many possible choices we selected five widely used time series analysis methods that did not require long runs of baseline data: the Centers for Disease Control and Prevention (CDC) Rapid Assessment Surveillance System (EARS) C2-Medium and C3-Ultra [19], the W2 variation [20] allowing for weekday/weekend trends, the moving F-statistic [21] and the exponential weighted moving average (EWMA) [22]. In the experiments we report on all these methods using a sliding 7 day baseline on a univariate measure of the document frequency counts on the topic of a specific disease in a specific country location. The results highlight the contribution online health news reports can make for system developers, methodologists and public health users.

## Methods
### Text mining system
BioCaster [4], an operational system since 2006, is a fully automated text mining system for monitoring global online media. The system's coverage is over 1700 Really Simple Syndication (RSS) feeds including sources from Google News, World Health Organization (WHO) outbreak reports, ProMED-mail, the European Media Monitor and other local and national news providers. The list of RSS news feeds is typically checked on a 60 minute cycle although this can be shortened as required.

The study period for the evaluation was chosen as the 366 days from 17[th] June 2008 to 17[th] June 2009. In order to generalize our conclusions, we tested five alerting algorithms across a diverse range of 14 countries and 11 infectious disease types. We used both commonly occurring seasonal diseases such as dengue and yellow fever as well as more sporadic diseases such as anthrax, plague and the novel A(H1N1)/A(H5N1) influenza types. There was no specific effort made at this stage to model parameters for diseases such as incubation periods or transmission rates.

As shown in Figure 1, the system adopts a high throughput pipeline approach, starting with computationally inexpensive tasks for filtering out the majority of news outside its case definition. This aims to minimize the time delay between download and notification to users.

At the beginning of the pipeline, raw news articles are downloaded and cleaned to remove noise such as metadata or links to other stories. The first stage of semantic understanding was to classify texts into relevant or non-relevant based on a case definition. Case definition guidelines were adapted from the WHO International Health Regulations (IHR) 2007 [1] annex 2 decision instrument. The IHR are a method for countries to assess whether to notify public health emergencies to WHO. These guidelines were used in our previously reported studies to hand tag a corpus of 1000 news documents [15] which were then used to train a naive Bayes classifier. F-scores for topic classification were reported in [23] as 0.93.

For relevant documents we automatically annotated entities of interest for 18 concept types based on the BioCaster ontology [11] including diseases, viruses, bacteria, people, locations, symptoms and organizations.

Entity recognition, grounding and event extraction are performed by a specially developed parser called Simple Rule Language (SRL) based on regular expressions manually defined over semantically defined entity classes. Disease-country relations for events of interest are of particular importance for this study but several other domain-based conceptual relations are also tagged.

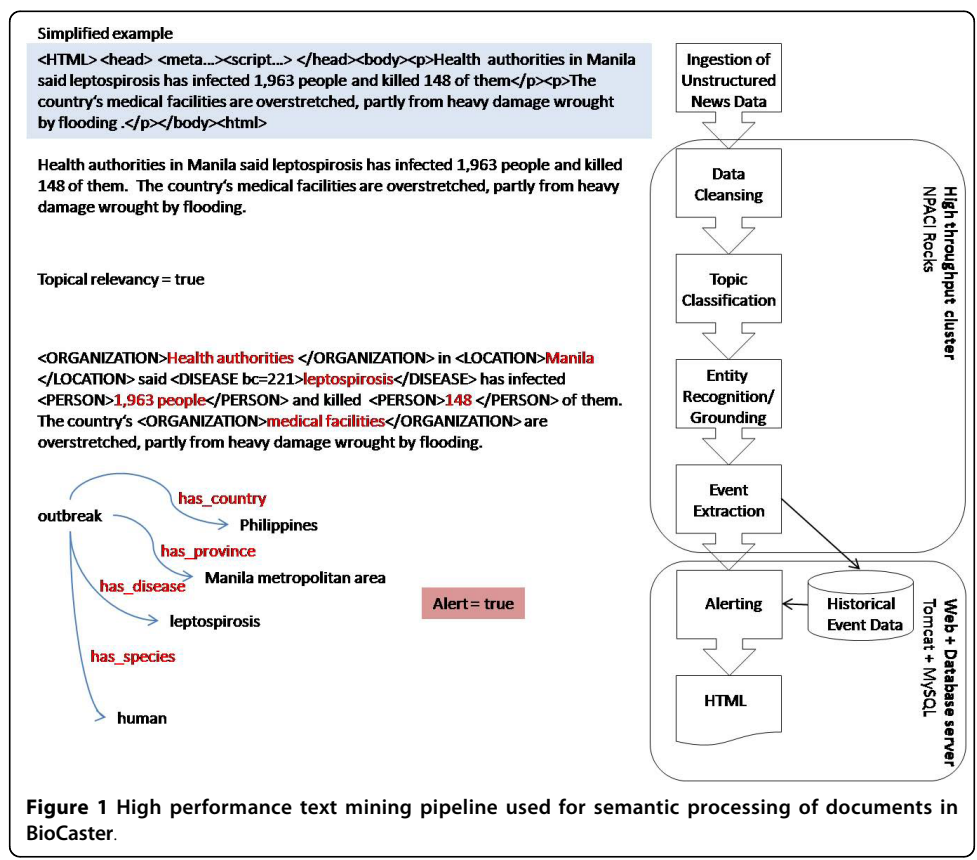

**Figure 1** High performance text mining pipeline used for semantic processing of documents in BioCaster.

An example of an SRL named entity rule is given below which would tag phrases such as *viral hemorrhagic outbreak* and *viral encephalitis outbreak* as entities of type *DISEASE*.

Ex 1. :- name(disease) { "viral" words(,1) "outbreak" }

More elegant and powerful rules can be made by employing word list sets:

Ex 2. :- name(disease) { list(%virus) "infection" }

which would tag any phrase in the word list set 'virus' followed by the literal 'infection' as an entity of type *DISEASE*. Other regular expression operators include skip words, orthographic matching, begins word and ends word. The following is an example of a template rule for identifying the main country in the outbreak event.

Ex 3. country(L) :- "situation" words(0,2) name(location, L) { list(@country) }

The body of the rule on the right hand side of the ':-' requires a match of a literal 'situation', followed by skipping up to 2 words, followed by a named entity of type 'location' containing a string in the 'country' word list. When the body of the rule matches the head expression, 'country(L)' is output where the variable L has been bound to the matched location entity in the body.

SRL rules have the advantage that they can be updated on-the-fly, e.g. when 'swine influenza' was officially re-designated as 'influenza A(H1N1)' without the need for retraining as in corpus-based approaches. They are also easily adapted to many languages and we have developed rule sets for diverse languages including Japanese, Thai, French and Vietnamese. SRL is freely available as a Google Code project [24] along with an introductory handbook and video.

The SRL rulebook for English was developed by hand based on a development corpus of over 1000 news articles drawn from a variety of international sources as well as a corpus of 5000 ProMED-mail reports published between December 1999 and July 2004. There was no overlap between the events in the development corpus and the evaluation corpus (see below). The English SRL rulebook used in this study contains approximately 110 rules, 12 major word list sets with 870 terms and over 2800 template rules.

SRL rules are used for entity grounding and event extraction. In grounding, disease and pathogen names are mapped to their root terms, i.e. the indexical terms in the BioCaster ontology. For example, the abbreviation 'WNV' becomes the root term 'West Nile Virus' using term equivalence relations encoded in the ontology. Multilingual equivalence relations in the ontology also allow us potentially to match events across languages.

Using SRL, it was straightforward to model simple inferred relationships. For example, country names are determined based not only on explicitly mentioned locations but also inferred from country adjectives (e.g. 'British officials' implies 'United Kingdom'), province mentions (e.g. 'Bagmati' implies 'Nepal'), major city mentions (e.g. 'Kisumu' implies 'Kenya') or organization names (e.g. 'HPA' implies 'United Kingdom'). An example of an inferred relation for diseases is the bacterium '*Legionella pneumophila*' causing 'Legionnaires' disease. In a news article with several grounded mentions of countries and diseases the choice of country is based on a heuristic context holding assumption, i.e. that whenever the country, disease or outbreak time changes then a new event is considered as being reported. We based our anomaly detection system on aggregated daily event counts.

Events are minimally required to identify the disease, country and time period (historical, recent past, present or hypothetical). The first event frame was chosen for each document, aiming to correspond to the main topic of the news report. To reduce double counting of duplicate reports, the document's URL (Universal Resource Locator) and the first 20 characters of its title are used. Since we aim for fully automatic alerts no manual verification of either news reports or country-disease relations was done.

### Gold standard data

As noted in the Background, we chose to base our gold standard on the reports from ProMED-mail, a freely available service provided through a network of experts. With its use of expert moderators, global coverage of infectious diseases and a staged review methodology we considered this to be a reasonably objective measure of alerting performance. The ProMED-mail archives were searched for documents containing the country and disease terms in the body or the title of the message and then each report was checked by hand by the author against the case definition. Reports that mentioned either suspected or confirmed cases were included. International travel events were recorded as being positive for both their source and target countries, e.g. a report of a case of a new type influenza traveling from Mexico to Canada would be considered positive for both Mexico and Canada. We regarded each ProMED-mail report collected in this way as an alert that something unusual might be happening.

Not all ProMED-mail postings were included in the gold standard. Filtering by hand removed a number of reports that fell outside of our case definition such as (a) aggregated summary reports at the country level that did not seem to arise from specifically mentioned events, (b) reports that primarily discussed the epidemiological or genetic characteristics of a disease agent, (c) reports that requested information, (d) reports that discussed non-outbreak events arising from an outbreak more than 3 months previously, (e) reports that primarily discussed control measures. The final collection of 18 test outbreaks is summarized in Table 1. In the same way, 5 additional data sets were collected as training data to calculate alerting thresholds for the detection models. These were: Chikungunya in India (11 ProMED-mail postings), Anthrax in Russia (10), Influenza in Mexico (20), Plague in USA (9) and Chikungunya in Malaysia (15).

Selection of the 18 test data sets was done in order to maximize geographic representation and the variety of diseases with the exception of influenza which continues to be of particular significance during the ongoing A(H1N1)/A(H5N1) outbreaks. We included one outbreak of an animal disease (Ebola Reston in the Philippines - data set 9) because of its potential zoonotic interest. Although Ebola Reston virus is not known to cause serious illness in humans its transmission, virulence and other epidemiological facts were poorly understood. We also introduced a negative outbreak (Yellow fever in Senegal - data set 16) to see how our test bed system and alerting algorithms would handle this. A further outbreak, Anthrax in the USA (data set 7), was also in fact negative but considered positive by the gold standard; ProMED-mail reports commented on the 2001 historical white powder scare outbreaks and new evidence concerning the person responsible. Despite the absence of victims within the time frame of interest we chose to consider this as a positive outbreak for alerting purposes due to its public health interest.

Table 1 Summary of the 18 data sets used in the study

|    | Condition       | Country         | Official language | Internet penetration (%) | ProMED postings[A] |
|----|-----------------|-----------------|-------------------|--------------------------|--------------------|
| 1  | Hand Foot Mouth | PR China        | zh                | 22.4                     | 9                  |
| 2  | Ebola           | Congo           | fr                | 1.8                      | 17                 |
| 3  | Yellow Fever    | Brazil          | pt                | 34.4                     | 28                 |
| 4  | Influenza       | USA             | en                | 73.2                     | 21                 |
| 5  | Cholera         | Iraq            | ar                | 1.0                      | 5                  |
| 6  | Chikungunya     | Singapore       | en                | 67.4                     | 8                  |
| 7  | Anthrax         | USA             | en                | 73.2                     | 15                 |
| 8  | Yellow Fever    | Argentina       | es                | 49.4                     | 5                  |
| 9  | Ebola Reston    | Philippines     | tl, en            | 21.5                     | 15                 |
| 10 | Influenza       | Egypt           | ar                | 12.9                     | 49                 |
| 11 | Plague          | USA             | en                | 73.2                     | 8                  |
| 12 | Dengue Fever    | Brazil          | pt                | 34.4                     | 27                 |
| 13 | Dengue Fever    | Indonesia       | id                | 10.5                     | 14                 |
| 14 | Measles         | United Kingdom  | en                | 71.8                     | 13                 |
| 15 | Chikungunya     | Malaysia        | ms                | 62.8                     | 15                 |
| 16 | Yellow Fever    | Senegal         | fr                | 6.1                      | 0                  |
| 17 | Influenza       | Indonesia       | id                | 10.5                     | 35                 |
| 18 | Influenza       | Bangladesh      | bn                | 0.3                      | 3                  |

Official language is given in ISO 3166-1 alpha 2; Internet penetration statistics are from Internet World Statistics March 31st 2009. [A]'ProMED postings' shows the total number of true positives for each outbreak data set over the test study period.

In the absence of ground truth data, our goal in these experiments is to test our system's ability to make a first identification of an outbreak topic from open media sources compared to ProMED-mail outbreak reports. Topics needed to define a specific event with a named disease, country and time. Several simplifying assumptions were adopted: (a) any system alert on the same topic 7 days prior to a ProMED-mail report was considered as a true alarm, (b) any system alert outside of this time frame was considered as a false alarm, and (c) ProMED-mail alerts were not required to be topically independent.

### Detection methods

The algorithms we use here are well documented in the disease aberration literature e.g. [19-22,25-28]. All aim to detect short term aberrations in the time series when news volume reporting for a country-disease relation is higher than expected. News document counts were sampled for each country-disease pair at regular intervals of time $t$ which was set to a single day. Test statistic values were then calculated independently for each data set and compared against threshold values to decide on whether an unusual movement had taken place. Threshold values and other model parameters were optimized manually for each algorithm on the five training data sets in order to maximize F1. If the threshold was exceeded, a binary alert was issued.

Below we provide details of the five algorithms used in the study.

(1) The test statistic for the Early Aberration Reporting System (EARS) C2 algorithm [19] is calculated using a 7 day sliding window baseline with a 2 day guard on the target day $t$ being assessed. The target value was calculated as:

$$S_t = \max(0, (C_t - (\mu_t + k\sigma_t))/\sigma_t)$$

where $C_t$ is the count on the target day, k is a constant, set in our simulations to 1, and $\mu_t$ and $\sigma_t$ are the mean and standard deviation of the counts during a series of past time instants which we refer to as the baseline window. $S_t$ captures the number (k) of standard deviations that the count exceeds the baseline mean, defaulting to 0 if it does not exceed the baseline mean.

(2) C3 [19] is based on a modified version of the Cumulative Summation (CUSUM) algorithm [27]. Unlike other CUSUM algorithms, C3 uses only the two previous observations and is calculated in a similar manner to C2 except that it is the sum of $S_t + S_{t-1} + S_{t-2}$. $S_{t-1}$ and $S_{t-2}$ are added if the counts on those days do not exceed the threshold of 3 standard deviations plus the mean on those days. C3 is designed to extend the sensitivity of C2 and will generate an alert at least as often as C2.

(3) W2 is a stratified version of C2 which takes into account the possibility of day of week effects on the data source. This simplified approach buckets days into weekday and weekend, counting only the last 7 weekdays within the baseline window. Alerting however can occur on both weekdays and weekend days.

(4) The F-statistic [21] is calculated as:

$$S_t = \sigma_t^s + \sigma_b^2$$

where $\sigma_t^2$ approximates the variance during the testing window and $\sigma_b^2$ approximates the variance during the baseline window. Both are calculated as follows:

$$\sigma_t^2 = \frac{1}{n_t} \sum_{test}^{n_t} (C_t - \mu_b)^2$$

$$\sigma_b^2 = \frac{1}{n_b} \sum_{test}^{n_b} (C_t - \mu_b)^2$$

(5) Finally, EWMA [22] provides for less weight to be given to days that are further from the test day. The smoothed counts are calculated as follows:

$$Y_1 = C_1$$
$$Y_t = \lambda C_t + (1-\lambda) Y_{t-1}$$

where $0<\lambda<1$ controls the amount of smoothing, i.e. the sensitivity to smaller or larger deviations. In our simulations, we found that the optimal level of $\lambda$ was 0.2. Then the test statistic is calculated as:

$$S_t = (Y_t - \mu_t) / [\sigma_t * (\lambda / (2-\lambda))^{1/2}]$$

where $\mu_t$ and $\sigma_t$ are the mean and standard deviation for the baseline window.

The moving window for the time series is illustrated in Figure 2 using document frequency counts from BioCaster for the topic pair A(H1N1) influenza and United States. All methods used a history window of length 7 and all methods were preceded by a purge on the data to remove single frequency counts which we had observed contained a large number of anomalies (see Discussion). No upper threshold was set on frequency counts. Alerts were generated if the algorithm test statistic exceeded a threshold, which was determined experimentally to find the optimal level for each model across all 5 training data sets and set at 0.3 (C3), 0.2 (C2), 0.2 (W2), 0.6 (F-statistic), and 2.0 (EWMA). We found that tuning the thresholds and other parameters has a significant effect on performance.

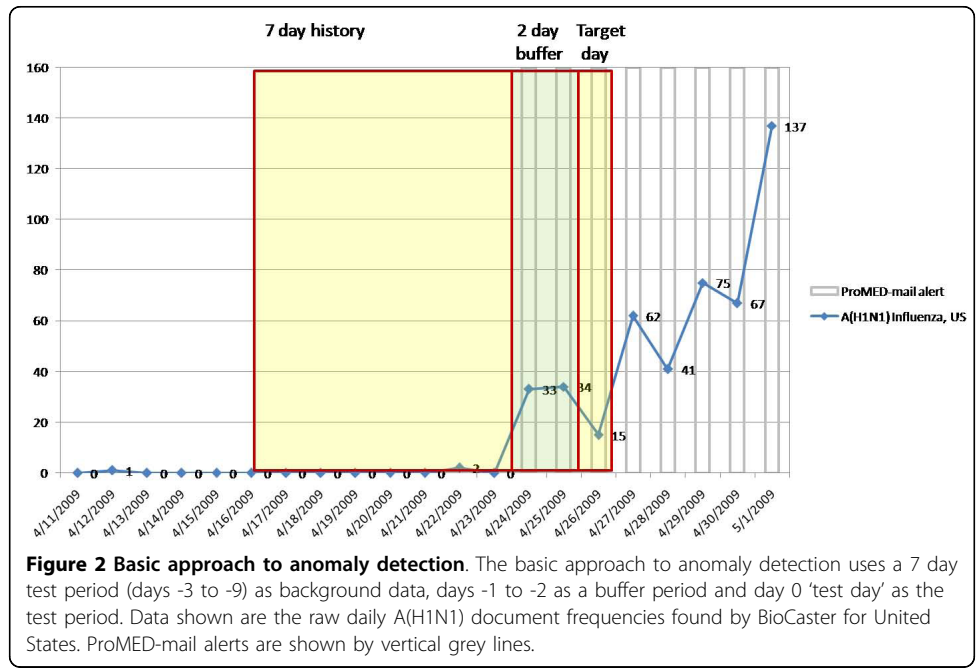

Figure 2 **Basic approach to anomaly detection**. The basic approach to anomaly detection uses a 7 day test period (days -3 to -9) as background data, days -1 to -2 as a buffer period and day 0 'test day' as the test period. Data shown are the raw daily A(H1N1) document frequencies found by BioCaster for United States. ProMED-mail alerts are shown by vertical grey lines.

Due to data sparseness we were in constant danger with standard deviation of division by zero so we implemented a minimum value for all models, finding an optimal level of F1 at a standard deviation value of 0.2. The minimum standard deviation interacts with the minimum frequency purge. For example with the threshold optimized and a zero count background for all 7 days, an alert is generated for C2 when the current daily count is 3 or more.

### Evaluation metrics

Our purpose in this evaluation was to obtain system alerts on or before the date of gold standard alerts. The period for a qualifying system alert on a country-disease topic was set as being 7 days prior to and including a qualifying ProMED-mail posting on the same topic (i.e. excluding postings that fell outside of the case definition which we discussed earlier). True positives were increased by 1 if there was any system alert that fell within the 7 day period. Multiple system alerts during this period did not count twice. False positives were increased by 1 for every system alert that fell outside of a 7 day qualifying alert period. False negatives were counted as the number of qualifying alert periods when there were no system alerts. True negatives were counted as the number of days outside of any qualifying alerting period when no system alert was given.

Evaluation metrics are then defined in the usual way as:

Sensitivity = TP / (TP + FN)
Specificity = TN / (TN + FP)
Positive Predictive Value (PPV) = TP / (TP + FP)
Negative Predictive Value (NPV) = TN / (TN + FN)

We also calculate the F1 score [29] which is the harmonic mean of sensitivity and PPV as:

$$F1 = 2 * (PPV * Sensitivity) / (PPV + Sensitivity)$$

## Results

### Aggregated accuracy

Aggregated performance for each detection method across all the 18 data sets is shown in Table 2. Combining results in this way helps to smooth out the differences in the alerting volume and onset type of outbreaks (spike or gradual slope) that we observe in individual data sets. The performance shown is the best that each model could achieve when its thresholds and other parameters were optimized for F1 [29], i.e. a single measure combining the mean of sensitivity and PPV that is commonly used in the text mining community. F1 can be interpreted as giving equal weight to the probability that a true alert will be found and the probability that a system alert will be a true alert. When considered from this perspective we find that W2 performs best in terms of F1 (0.63). This result is led by the strong PPV which shows that W2 would generate the fewest number of false alarms - important when the cost of further investigation is potentially very high.

We notice also an indication of a day of week effect in the F1 results between C2 (0.629) and W2 (0.633) which uses a stratified approach based on weekend and weekday counts. This in turn is supported by analysing mean news volumes aggregated across all 18 data sets. The mean number of documents detected by the text mining system per day on each of the country-disease topics was 1.37 for weekdays (n = 6454) and 0.49 for Saturdays and Sundays (n = 928). This clearly indicates a bias in the news for weekdays.

Further analysis of the aggregated alerts from all models and of the aggregated postings from ProMED-mail shown in Figure 3 reveals a more complex situation than we had at first imagined. Posting counts for ProMED-mail on dengue topics indicated a strong bias towards reporting on Monday or Tuesday (left side graph) whereas dengue news volumes do not (data not shown). When the dengue data sets are excluded (right side graph) the alerting curves for the 5 models and ProMED-mail postings fall into broad agreement. Alerts tend to peak on Thursday, falling off greatly from Saturday through Monday.

**Table 2 Aggregated evaluation metrics across the 18 data sets**

|  | C3 | C2 | W2 | F-Statistic | EWMA |
|---|---|---|---|---|---|
| Sensitivity | 0.78 | 0.73 | 0.72 | 0.80 | 0.73 |
|  | (0.74-0.82) | (0.68-0.78) | (0.67-0.77) | (0.77-0.83) | (0.68-0.78) |
| Specificity | 0.95 | 0.97 | 0.97 | 0.91 | 0.95 |
|  | (0.94-0.95) | (0.96-0.97) | (0.97-0.97) | (0.90-0.91) | (0.94-0.96) |
| PPV | 0.49 | 0.55 | 0.56 | 0.45 | 0.47 |
|  | (0.44-0.53) | (0.50-0.60) | (0.51-0.61) | (0.42-0.48) | (0.43-0.52) |
| NPV | 0.99 | 0.99 | 0.99 | 0.98 | 0.98 |
|  | (0.98-0.99) | (0.98-0.99) | (0.98-0.99) | (0.98-0.98) | (0.98-0.99) |
| Alarms/100 days[A] | 7.65 | 5.98 | 5.57 | 13.77 | 7.85 |
| F1 | 0.60 | 0.63 | 0.63 | 0.58 | 0.58 |

[A] This compares to a mean of 4.36 alerts per 100 days for ProMED-mail on the 18 data sets. Figures in parentheses show 95% CI.

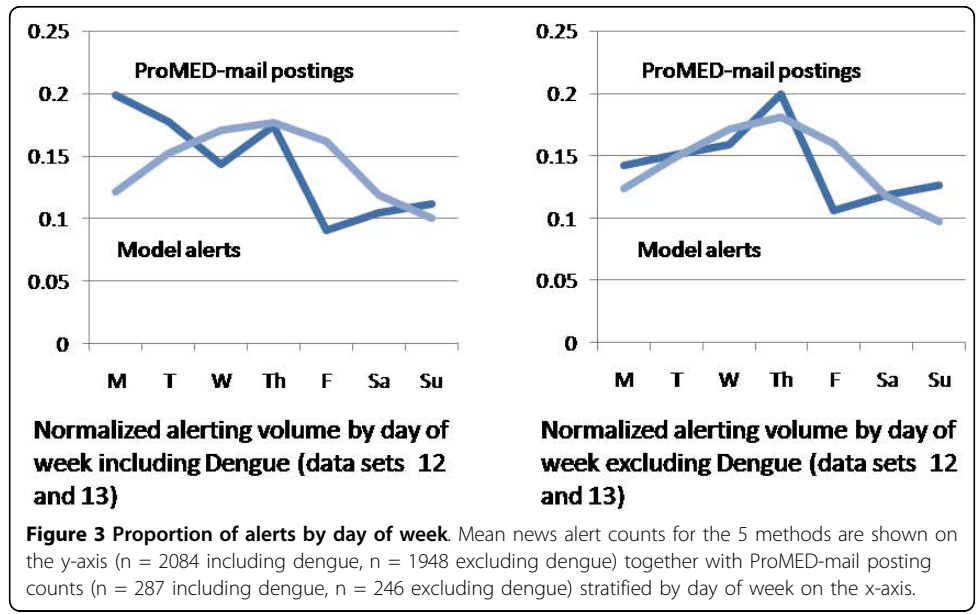

**Figure 3 Proportion of alerts by day of week**. Mean news alert counts for the 5 methods are shown on the y-axis (n = 2084 including dengue, n = 1948 excluding dengue) together with ProMED-mail posting counts (n = 287 including dengue, n = 246 excluding dengue) stratified by day of week on the x-axis.

As shown in Figure 4, not counting the special case of data set 16 for which there were no true positives, W2 sensitivity ranged from a low of 0.24 for Dengue fever in Brazil (data set 12) to a high of 1.0 for Influenza in Bangladesh (data set 18). PPV for W2 ranged from a low of 0.21 for Yellow fever in Argentina (data set 8) to highs of 1.0 for Dengue fever in Brazil (data set 12) and 0.94 Influenza in Egypt (data set 10).

Several other interesting trends are shown by the results in Figure 4. As expected the F-statistic and C3 both have strong sensitivity (0.80, 0.78) compared to C2 and W2 (0.73, 0.72). However both the F-statistic and EWMA suffered relatively low PPV (0.45, 0.47). These are likewise reflected in the mean number of alerts/100 days. Of the five methods C2 and W2 had the closest mean number of alerts/100 days to the gold standard baseline for ProMED-mail (4.36). NPV remains uniformly high due to the strong bias in non-alerting days across the data sets.

### Outbreak characteristics

Consistent with other data sources, e.g. [22], our results show that there is a tendency for the longer runs of news reports to adversely affect overall detection performance. As shown by cross referencing Figures 4 and 5, spike reports such as those we observed for Cholera in Iraq (data set 5) and Measles in the UK (data set 14) tended to do comparatively well compared to gradual slope news events such as Influenza in the USA (data set 4). In this respect it is interesting to contrast the gradual slope for the Influenza A (H1N1) pandemic (data set 4) with the many short spikes that we observed for Influenza A(H5N1) in Egypt (data set 10). Performance by models on the former were uniformly disappointing due to high numbers of false negatives whilst for the latter were among the highest of any data set. The initial reason for this would appear to be that the aberration detection algorithms are designed to detect outbreaks on the upward trajectory of the curve but not on a downwards path when the outbreak numbers are decaying. This explanation though is not completely satisfactory and hides two other important factors - location granularity and reporting policy - which we discuss below.

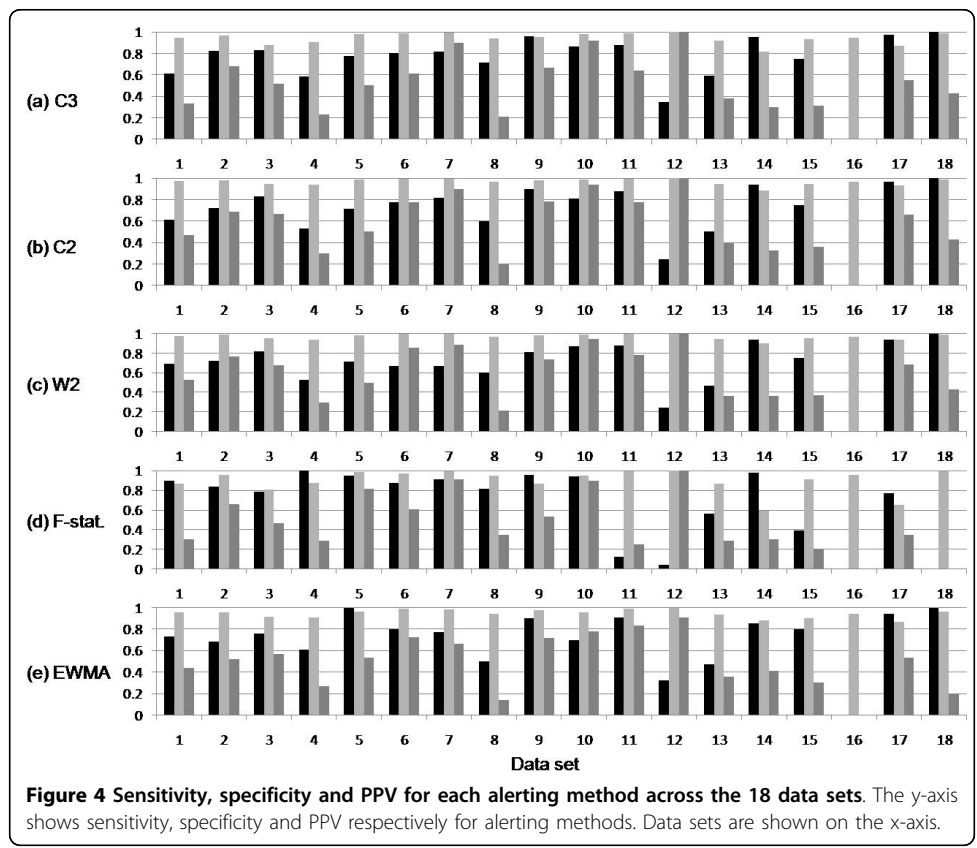

Figure 4 **Sensitivity, specificity and PPV for each alerting method across the 18 data sets**. The y-axis shows sensitivity, specificity and PPV respectively for alerting methods. Data sets are shown on the x-axis.

The negative outbreak of Yellow Fever in Senegal (data set 16) proved challenging for all of the algorithms. Several small spikes of mentions were found in the news that related Senegal to Yellow Fever; however the relationship was actually a false one during the observation period with 'Senegal' being the location of a WHO reference laboratory whilst the actual outbreak took place elsewhere in other parts of Africa. This illustrates the necessity for GHIS to have a precise understanding of the relationship between disease and location in text.

**Missing data**

Sinks are sudden drops in the daily document count to zero and then a sudden rise again. Like the one we observed for Ebola in Congo (Figure 5) these are often traceable to day of week effects. For example, the drop on $27^{th}$ and $28^{th}$ December 2008 and on the $3^{rd}$ and $4^{th}$ of January 2009 both occur on weekend days. Similarly, the drops for Influenza reports in the USA (Figure 2) on the $26^{th}$ April, $3^{rd}$ and $10^{th}$ May 2009 all occurred on Sundays. This raises the question for how to compensate for this effect in future work.

**Data filtering**

Our decision to purge the daily counts of low frequency reports appeared to pay off. Overall F1 performance is increased by approximately 1 to 1.5 (data not shown) when we removed document counts of 1 or 2 (but not 3 or higher) from the system. On examination we found that singleton reports can often be reassurances that the situation is under control or simple updates on preventative measures that are taking place.

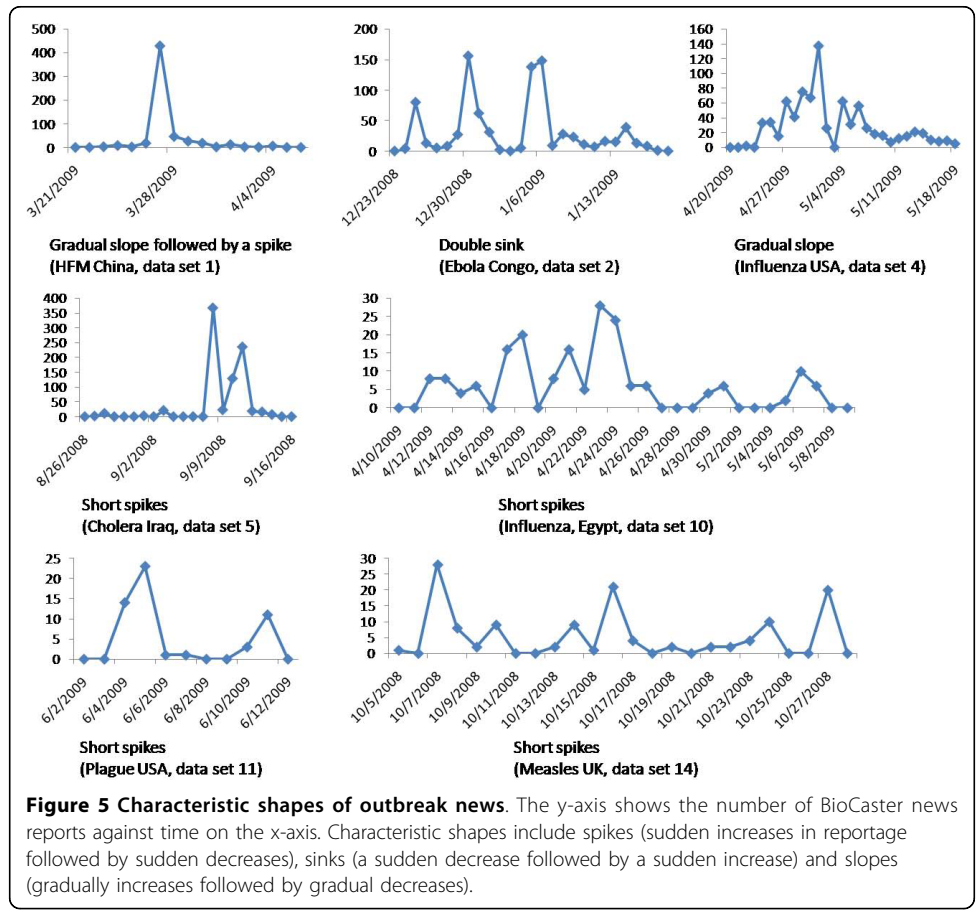

**Figure 5 Characteristic shapes of outbreak news**. The y-axis shows the number of BioCaster news reports against time on the x-axis. Characteristic shapes include spikes (sudden increases in reportage followed by sudden decreases), sinks (a sudden decrease followed by a sudden increase) and slopes (gradually increases followed by gradual decreases).

## Discussion

Whilst the aggregated results indicate the optimal choice of alerting algorithm, a more fulfilling understanding emerges from a drill down inspection into the data.

### Extended outbreaks

As we noted earlier, all of the detection methods we used look for alerts when news report coverage is increasing. On the downward slope of gradual outbreaks there are normally no alerts generated by the models. We observed that in many cases though that ProMED-mail continued to send postings during such periods and the true positive rate for most algorithms decreased, impacting on PPV. There are two points to observe here:

> (a) Country level modeling hampered us during ongoing outbreak monitoring as seen in gradual slope reporting. True alerts are often generated whenever a first case of a disease appears in a province or other sub-country unit of health jurisdiction. Without considering this level of granularity the automatic alerting systems will be blind to these types of aberrations, i.e. all news will appear part of a single distribution. This raises the question, would a lower granularity model of location perform better?
> (b) ProMED-mail reporting policy seemed to adjust between the outbreak detection and outbreak tracking stages of some epidemics. For example, in an outbreak tracking mode, distinctions between suspected and confirmed cases or between living cases and fatalities become relatively more important. During the A(H1N1) Influenza outbreak in the USA (data set 4) we observed that ProMED-mail included reports that focused on providing information to assess the potential longer term impact of the agent. Similarly, dengue postings from ProMED-mail are sometimes summaries of situations that could have started weeks or months before and are still ongoing. This weakens one assumption in our gold standard which is that initial ProMED-mail postings will be generated within a week of news (if any) on the outbreak (see the section on Gold standard data). Any alerts the system makes outside of the 7 day window, which may be valid, are counted as negatives. This may explain why the results for dengue in data sets 12 and 13 have comparatively low sensitivity.

Extended gradual outbreaks on the ground cause difficulties in other ways too, as we saw in the Yellow Fever outbreak in Brazil (data set 3). In cases like this, the true time frame for outbreak alerts appears to have a large margin for uncertainty. On analysis we found that the system alerts generated for Yellow Fever between August 6th-8th 2008 appeared to be reasonable according to our case definition but there was no corresponding report on ProMED-mail for South America. The spike was generated by reports from European Media Monitor alerts, the Canadian Press, the CDC and healthnews.com on warnings to travelers about Yellow Fever in Brazil. Many of these could be traced back to a Brazilian health ministry alert to reinforce recommendations for Yellow Fever vaccination among travelers to areas where the disease regularly occurs. It is difficult to know for sure if this was connected to an ongoing outbreak, the subsequent outbreak alerts from November onwards or to the earlier outbreak alerts in July. The cluster of reports though from at least three official sources does indicate heightened awareness of the situation in the region.

### Non-alerting events

Other spike reports from the system are more clearly false positives and show areas where we must improve the level of textual understanding. A Yellow Fever spike for Brazil on September 4th came from a story about the procurement of vaccines by Guyana from Brazil. Similarly a spike of reports on Dengue for Indonesia occurred around 7th May 2009 when the Jakarta Post newspaper seems to have run several stories on Dengue fever at the peak of the regular Dengue season. This accompanied the first fatality from Dengue in the capital and a campaign by the Governor of Jakarta to improve environmental cleanliness. The reported stories though all seem to have come against a background of recent unusually high numbers of Dengue cases. In this case it complicated the task of differentiating between a prevention campaign and an outbreak.

Differential classification of borderline media reports such as those that focus mainly on control campaigns has been strengthened in our previous studies [15] by trying to focus the initial stage of topic classification on more dominant concept roles in the outbreak event such as infected individuals or groups or the status of chemicals as drugs. This line of research - i.e. focusing on the status of victims in the report - could be strengthened further for example by excluding or down weighting sentences that do not contain the victim role or related concepts such as their health status or the social reactions of at risk groups. Further improvements might be seen downstream at the alerting stage by improving extending the baseline period for the background model.

### Border regions

An interesting case study we observed occurred around 6th January 2009 when a spike of news showed activity for Yellow Fever in Brazil several days in advance of a ProMED-mail alert. Local English language reports appeared about the deaths of two jungle workers in a border region between Brazil, Argentina and Paraguay occurring simultaneously with dead monkeys. At this stage sanitary authorities in the area were already conducting tests. ProMED-mail reported an alert under Argentina at this time and under Brazil a few days later. This raises the need to consider alerting at the multi-country level when provinces in neighbouring countries have cases.

Confusion arose within our system with a report on 30th March and 2nd June 2009 when ProMED-mail listed North Dinajpur in "West Bengal" as being the site of a bird cull to prevent A(H5N1) influenza. The Dinajpur area straddles India and neighboring Bangladesh. In this case the BioCaster system did not assign significance to "North" as part of the official name which caused the locale to be registered as Bangladesh in our system.

### Reporting bias

As we noted in the introduction there may be an inherent bias in how likely an event on a given topic is to be reported. These are however possibly the most difficult effects to compensate for and were not quantified here as we felt an in-depth treatment was beyond the scope of the present study. Surveys such as [30] shed light on this area by showing that news organizations tend to cover some events more fully than others, reflecting events that impact on the readership audience or which are highly disruptive or unexpected. The number of local news organizations differs among countries as does the allocation of reporters from international news agencies. The latter tends to be biased according to those that have cultural and economic links to the home country lending increased coverage to countries such as the United States.

The study we present here for system alerts on English language news needs to be extended in the future to qualitatively compare alerting accuracy in countries where the main media language was a major world language that was not English (e.g. Brazil, Russia, China). Anecdotally, we observed for example that the Dengue epidemic that occurred in Ceara Brazil around July-September 2008 went unreported by our system perhaps because at that time we did not have a Portuguese capability. The early stages of the 2009 A(H1N1) influenza pandemic in Mexico also went unreported by our system until April 21st. Again, this is because we lacked Spanish language coverage. Both Portuguese and Spanish have now been added.

By comparing Table 1 to Figure 4, it is worth noting that we could not find any obvious linkage between alerting F1 value and the Internet penetration in each of the countries studied. This most likely reflects a bias in the BioCaster media coverage at the time of the study towards reports from international agencies for country-level modeling. We can speculate that the relationship between Internet penetration and alerting accuracy might become apparent at province level modeling or when collecting health intelligence data from blogs. Coverage of localized sources also becomes more critical for other proxy methods such as measuring user query activity [31].

## Conclusions

Web-based health news has the potential to enhance manual analytical methods by increasing throughput, timeliness and detection rates. The relationship between news counts and actual numbers of cases is a complex one and cannot at this stage be assumed as simply proportional due to the factors we have discussed. Raw news counts tend to reflect the degree of concern felt by the reader population which may not always be the population at risk. Systematic evaluation of health news aberrations are therefore necessary (a) to help improve our understanding of this relationship, (b) to distinguish the best performing algorithms and features and, (c) to highlight challenge areas ahead for textual understanding. Increased transparency in evaluation will likewise benefit both users and system developers.

In this article we have shown the performance of five widely used detection algorithms on an open baseline. Through BioCaster's data, we have demonstrated global surveillance capability but one with interesting variations in performance. When maximizing sensitivity and PPV, the W2 algorithm outperformed four other methods, but we observed wide variations among models across individual epidemics due to complex underlying limitations and assumptions such as the geographic level on which we model data, reporting bias at the country and language level and the case definition of an alert. Future research is needed to help decide on the best denominators for normalizing news counts such as population numbers or news volume stratified by regions.

In the future we intend to explore the underlying limitations of performing geotemporal classification at lower levels of granularity. The challenge here is that this will exacerbate data sparseness and lead to a more fragmented distribution of features. In order to extend the volume of features we have available to us we have already taken steps to increase the number of news sources from nearly 1700 to over 20,000 by outsourcing data gathering to a commercial news aggregation company.

Given the data characteristics we have observed so far, a robust solution to detecting the initial very small numbers of news reports at the front of an epidemic wave is unlikely to come only from counting disease-country relations. Just as with human analysts, what is required is rather to employ a deeper understanding of other relations inside and outside the news report (e.g. public health, social and economic responses to events) as well as integration with other signal sources (e.g. climatic data) to make sense of more detailed contextual signals. This will also form part of our future investigation.

### Acknowledgements
I would like to express my gratitude to the many people who have helped in the development of the BioCaster system and ontology: Son Doan (Vanderbilt University Medical Center), Reiko Matsuda Goodwin (Fordham University), Ai Kawazoe (Tsuda College), Mika Shigematsu and Kiyosu Taniguchi (National Institute of Infectious Diseases), Koichi Takeuchi (Okayama University), Mike Conway (University of Pittsburgh), John McCray (University of Bielefeld), Dinh Dien and Ngo Quoc-Hung (Vietnam National University), Asanee Kawtrakul (NECTEC and Kasetsart University), Yoshio Tateno (National Institute of Genetics) and Roberto Barrero (Murdoch University).
Financial support: Japan Science and Technology Agency (JST)'s PRESTO fund.

**Author details**
[1]National Institute of Informatics, 2-1-2 Hitotsubashi, Chiyoda-ku, Tokyo 101-8430, Japan. [2]PRESTO, Japan Science and Technology Corporation, 2-1-2 Hitotsubashi, Chiyoda-ku, Tokyo 101-8430, Japan.

**Authors' contributions**
NC conceived of the study, carried out the experiments and data analysis as well as manuscript preparation.

**Competing interests**
The author declares that they have no competing interests.

**References**
1. Gostin LO: **International infectious disease law - Revision of the World Health Organization's International Health Regulations.** *Journal of the American Medical Association* 2004, **291(21)**:2623-2627.
2. Paquet C, Coulombier D, Kaiser R, Ciotti M: **Epidemic intelligence: a new framework for strengthening disease surveillance in Europe.** *Euro Surveillance* 2006, **11(12)**.
3. Keller M, Blench M, Tolentino H, Freifeld CC, Mandl KD, Mawudeku A, Eysenbach G, Brownstein JS: **Use of unstructured event-based reports for global infectious disease surveillance.** *Emerging Infectious Diseases* 2009, **15(5)**:689-695.
4. Collier N, Doan S, Kawazoe A, Goodwin RM, Conway M, Tateno Y, Ngo QH, Dien D, Kawtrakul A, Takeuchi K, Shigematsu S, Taniguchi K: **BioCaster: detecting public health rumors with a Web-based text mining system.** *Bioinformatics* 2008, **24(24)**:2940-2941.
5. Freifeld CC, Mandl KD, Reis BY, Brownstein JS: **HealthMap: Global Infectious Disease Monitoring through Automated Classification and Visualization of Internet Media Reports.** *Journal of the American Medical Informatics Association* 2008, **15(2)**:150-157.
6. Wilson J: **Argus: A Global Detection and Tracking System for Biological Events.** *Advances in Disease Surveillance* 2007, **4**:21.
7. Steinberger R, Flavio F, Goot van der E, Best C, von Etter P, Yangarber R: **Text Mining from the Web for Medical Intelligence.** *Mining Massive Data Sets for Security* IOS Press, Amsterdam, The NetherlandsFogelman-Soulié F, Perrotta D, Piskorski J, Steinberger R 2008, 295-310.
8. Kass-Hout T, di Tada N: **International System for Total Early Disease Detection (InSTEDD) Platform.** *Advances in Disease Surveillance* 2008, **5(2)**:108.
9. Heffernan R, Mostashari F, Das D, Karpati A, Kulldorff M, Weiss D: **Syndromic surveillance in public health practice: The New York City emergency department system.** *Emerging Infectious Diseases* 2004, **10**:858-864.
10. Eysenbach G: **SARS and Population Health Technology.** *Journal of Medical Internet Research* 2003, **5(2)**:e14.
11. Collier N, Kawazoe A, Jin L, Shigematsu M, Dien D, Barrero R, Takeuchi K, Kawtrakul A: **A multilingual ontology for infectious disease surveillance: rationale, design and challenges.** *Language Resources and Evaluation* 2007, 405-13 [http://biocaster.org].
12. Wilson JM, Polyak MG, Blake JW, Collmann J: **A heuristic indication and warning staging model for detection and assessment of biological events.** *Journal of the American Medical Informatics Association* 2008, **15**:158-171.
13. Hartley DM, Nelson NP, Walters R, Arthur R, Yangarber R, Madoff L, Linge JP, Mawudeku A, Collier N, Brownstein JS, Thinus G, Lightfoot N: **The Landscape of International Event-based Biosurveillance.** *Emerging Health Threats Journal* 2010.
14. Mawudeku A, Blench M: **Global Public Health Intelligence Network (GPHIN).** *In Proceedings of the 7th Conference of the Association for Machine Translation in the Americas: August 2006, Cambridge, MA, USA* .
15. Doan S, Kawazoe A, Conway M, Collier N: **Towards role-based filtering of disease outbreak reports.** *Journal of Biomedical Informatics* Elsevier 2008.
16. Kawazoe A, Jin L, Shigematsu M, Barerro R, Taniguchi K, Collier N: **The development of a schema for the annotation of terms in the BioCaster disease detection/tracking system.** *In Proceedings of the International Workshop on Biomedical Ontology in Action (KR-MED 2006): November 2006; Baltimore, Maryland, USA* Bodenreider O 2006, 77-85.
17. Madoff L: **ProMED-mail: An early warning system for emerging diseases.** *Clin Infect Dis* 2004, **39(2)**:227-232.
18. Woodall JP: **Global Surveillance of Emerging Diseases: the ProMEDmail Perspective.** *Cad Saude Publica* 2001, **17(Suppl)**:147-154.
19. Hutwagner L, Thompson W, Seeman MG, Treadwell T: **The Bioterrorism Preparedness and Response Early Aberration Reporting System (EARS).** *Journal of Urban Health: Bulletin of the New York Academy of Medicine* 2003, **80((2) suppl 1)**:i89-i96.
20. Copeland J, Rainisch G, Tokars J, Burkom H, Grady N, English R: **Syndromic prediction power: comparing covariates and baselines.** *Advances in Disease Surveillance* 2007, **2**:46.
21. Burkom HS: **Accessible Alerting Algorithms for Biosurveillance.** *National Syndromic Surveillance Conference* 2005.
22. Jackson ML, Baer A, Painter I, Duchin J: **A simulation study comparing aberration detection algorithms for syndromic surveillance.** *Medical Informatics and Decision Making* 2007, **7(6)**, BMC, DOI: 10.1186/1472-6947-7-6.
23. Conway M, Doan S, Kawazoe A, Collier N: **Classifying disease outbreak reports using n-grams and semantic features.** *Journal of Medical Informatics* 2009.
24. **Simple Rule Language Editor.** [http://code.google.com/p/srl-editor/].
25. Murphy SP, Burkom H: **Recombinant Temporal Aberration Detection Algorithms for Enhanced Biosurveillance.** *Journal of the American Medical Informatics Association* 2008, **15(1)**:77-86.
26. Hutwagner L, Browne T, Seeman GM, Fleischauer AT: **Comparing aberration detection methods with simulated data.** *Emerging Infectious Diseases* 2005, **11**:314-316.
27. Basseville M, Nikiforov I: **Detection of abrupt changes: Theory and Application.** *Prentice Hall* 1993.
28. Tokars JI, Burkom H, Xing J, English R, Bloom S, Cox K, Pavlin J: **Enhancing Time-Series Detection Algorithms for Automated Biosurveillance.** *Emerging Infectious Diseases* 2009, **15(4)**:533-539.

29. van Rijsbergen CJ: **Information Retrieval.** *Butterworth* 1979.
30. Wu HD: **Systemic Determinants of International News Coverage: A Comparison of 38 Countries.** *The Journal of Communication* 2006, **50(2)**:110-130.
31. Ginsberg J, Mohebbi MH, Patel RS, Brammer L, Smolinski MS, Brilliant L: **Detecting influenza epidemics using search engine query data.** *Nature* 2008, **457**:1012-1014.